\documentclass[letterpaper]{article} 
\usepackage{aaai25}  
\usepackage{times}  
\usepackage{helvet}  
\usepackage{courier}  
\usepackage[hyphens]{url}  
\usepackage{graphicx} 
\usepackage{natbib}  
\usepackage{caption} 
\usepackage{algorithm}
\usepackage{latexsym}
\usepackage{multirow}
\usepackage{amsmath}
\usepackage{algpseudocode}
\usepackage{latexsym}
\usepackage{mathletters}
\usepackage{float}
\usepackage[T1]{fontenc}
\usepackage{enumitem}
\usepackage{amssymb}
\usepackage{mathtools}
\usepackage{amsthm}
\usepackage{amsfonts}
\usepackage{dsfont}

\usepackage{caption}
\usepackage{subcaption}
\usepackage{eurosym,rotating,booktabs}

\usepackage[capitalize,noabbrev]{cleveref}
\usepackage{devanagari}
\usepackage[english]{babel} 
\babelprovide[import]{hindi}
\babelprovide[import]{sanskrit}
\theoremstyle{plain}

\theoremstyle{definition}

\theoremstyle{remark}

\title{Enhancing Entertainment Translation for Indian Languages using Adaptive Context, Style and LLMs}
\author{
    Pratik Rakesh Singh, Mohammadi Zaki and Pankaj Wasnik
}
\affiliations{
    Media Analysis Group, Sony Research India, Bangalore\\
    \{pratik.singh, mohammadi.zaki, pankaj.wasnik\}@sony.com
}

\usepackage{bibentry}

\usepackage{newfloat}
\usepackage{listings}
\DeclareCaptionStyle{ruled}{labelfont=normalfont,labelsep=colon,strut=off} 

\begin{document}

\maketitle

\begin{abstract}
We address the challenging task of neural machine translation (NMT) in the entertainment domain, where the objective is to automatically translate a given dialogue from a source language content to a target language. This task has various applications, particularly in automatic dubbing, subtitling, and other content localization tasks, enabling source content to reach a wider audience. Traditional NMT systems typically translate individual sentences in isolation, without facilitating knowledge transfer of crucial elements such as the \textit{context} and \textit{style} from previously encountered sentences. In this work, we emphasize the significance of these fundamental aspects in producing pertinent and captivating translations. We demonstrate their significance through several examples and propose a novel framework for entertainment translation, which, to our knowledge, is the first of its kind. Furthermore, we introduce an algorithm to estimate the context and style of the current \textit{session} and use these estimations to generate a \textit{prompt} that guides a Large Language Model (LLM) to generate high-quality translations. Our method is both language and LLM-agnostic, making it a general-purpose tool. We demonstrate the effectiveness of our algorithm through various numerical studies and observe significant improvement in the COMET scores over various state-of-the-art LLMs. Moreover, our proposed method consistently outperforms baseline LLMs in terms of win-ratio.
\end{abstract}

\section{Introduction}\label{sec:introduction}

Recent advancements in neural machine translation (NMT) have become increasingly important in the entertainment industry for automatic content localization. These advancements have addressed some limitations of entertainment translation by incorporating contextual understanding and cultural nuances into translations \cite{yao2024benchmarkingllmbasedmachinetranslation, matusov-etal-2019-customizing, vincent2024casestudycontextualmachine}. 

In entertainment content, where dialogues often depend on prior interactions to convey a scene's meaning and emotion effectively, context-aware translation plays a vital role \cite{context2, context1, vincent-etal-2024-reference, agrawal-etal-2023-context}. Incorporating the broader dialogue or narrative context, rather than translating sentences in isolation, is crucial to ensure accurate and emotionally relevant translations \cite{creative-subtitling}.

\begin{figure}
        \centering
        \includegraphics[width=.45\textwidth]{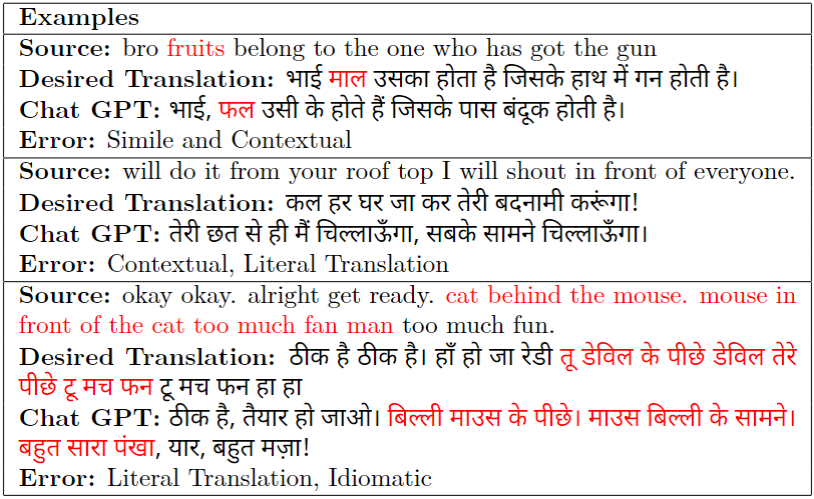}
        \caption{Examples of common mistakes made by NMT systems while translating entertainment domain text.}
        \label{fig:motive}
    \end{figure}

On the other hand, entertainment translation also needs a culturally adaptable system to address the challenge of cultural unawareness \cite{etchegoyhen-etal-2014-machine, yao2024benchmarkingllmbasedmachinetranslation}. Such systems should integrate cultural context for localization to ensure translations are suitable for the intended audience. They should go beyond literal translations, modifying idiomatic expressions, jokes, and cultural references to align with the audience's customs and values, thereby enhancing the relevance of the translated content\cite{gupta2019problemsautomatingtranslationmovietv,Li_Chen_Yuan_Wu_Yang_Tao_Xiao_2024}. In Figure \ref{fig:motive} we show some examples of common mistakes made by NMT systems when translating entertainment content. In Example 1, 'fruits' idiomatically refers to 'reward,' but ChatGPT's literal translation misses this. In Example 2, the desired translation is culturally more creative, aligning with native Hindi speakers by conveying “I will badmouth you by knocking door to door.” In Example 3, the desired translation uses idiomatic language effectively, unlike ChatGPT's literal approach.

In this paper, we address the challenging task of entertainment translation, where we are given a sequence of source sentences from the entertainment domain without any additional information about the timestamp, speaker ID, or context, and our task is to translate these sentences into dialogues in the target language. The challenge lies in preserving the context, mood, and style of the original content while also incorporating creativity and considering regional dialects, idioms, and other linguistic nuances \cite{gupta2019problemsautomatingtranslationmovietv}. The importance of our study is underscored by the need to produce translations that are not only accurate but also engaging for the target audience.

In particular, we treat the entertainment translation task as a sequential process to extract time-dependent contextual information by dividing the input text into a series of sessions. We primarily employ context-retrieval and domain adaptation to facilitate in-context learning of Large Language Models to extract both the style, representing the cultural nuances and temporal context from these \textit{sessions}. We can then use this characteristic information to generate culturally enriched translations. In addition, our proposed methodology does not need auxiliary information such as speaker information, timestamps, and conversation mood, making it generalized and applicable in a wide range of applications. Our key contributions can be summarized as follows:

\begin{itemize}
    \item We proposed an algorithm (Alg.\ref{alg:main pipeline]}), which we call Context And Style Aware Translation (CASAT). It incorporates context and style awareness, enhancing the input prompt and enabling LLM to produce culturally relevant translations.
    \item Proposed methodology is language and LLM-agnostic. further, it does not rely on dialogue timestamps, speaker identification, etc., making it a versatile approach.
    \item We proposed Context retrieval–Advanced RAG module to extract a precise and relevant context from entertainment content such as a movie or series episode.
    \item We proposed a Domain Adaptation Module to provide a cultural understanding of input to LLMs.  
\end{itemize}

\section{Background and Motivation}\label{sec:background}
In this section, we provide a review of some of the major research works in the field of machine translation as well as applications of LLMs in NMT.

NMT was introduced in the seminal works of \cite{Bahdanau1, Bahdanau2}, who used basic encoder-decoder architectures and RNNs, respectively, for the NMT task. These techniques were superseded by attention-based mechanisms introduced in \cite{attention-lstm, Google}. With the advent of Transformers in \cite{transformers}, the attention computation became massively parallelized, increasing the speed and efficiency of modern NMT systems.

\noindent{\bfseries{LLMs for NMT:}} In the last couple of years, LLMs have caused a major shift in the way AI research is carried out \cite{brown2020language}. The translation task has become a goto application of the LLMs since their advent \cite{lyu-etal-2024-paradigm}. A comprehensive review of machine translation using LLMs can be found in \cite{cai2024surveymixtureexperts}.

\noindent{\bfseries{Entertainment Translation:}} Most of the previously presented research on entertainment domain translation focuses primarily on subtitling and segmentation \cite{vincent-etal-2024-reference, karakanta-etal-2022-post, vincent2024casestudycontextualmachine,matusov-etal-2019-customizing, etchegoyhen-etal-2014-machine}. These works depend on additional information like timestamps and speaker details from the input text. However, timestamp information may not always be present or could be incorrect, leading to ambiguity or distortions in the temporal context, making entertainment translation more challenging \cite{gaido2024sbaameliminatingtranscriptdependency}.

\noindent{\bfseries{Use of contextual information for NMT:}} In recent years, the importance of (correct) context in the translation task has been studied and highlighted \cite{voita-etal-2019-good} for document-level translations \cite{context1}. However, these approaches do not perform consistently while dealing with overly large contexts or complicated scenarios \cite{context2}, as is usually the case in the entertainment domain.

\noindent{\bfseries{LLMs for Creative Translations and Style Transfer:}} Use of LLMs to induce creativity can be accomplished to a certain extent using prompt engineering techniques \cite{promptingllm}. In addition, advanced retrieval-based techniques \cite{agrawal-etal-2023-context, Reheman_Zhou_Luo_Yang_Xiao_Zhu_2023, glass2022re2gretrievererankgenerate} can be used to generate context from a given text and be used to provide necessary information for the desired translations. On the other hand, recent work on style transfer \cite{tao2024catllm} introduces a Domain Adaptation Module to copy the style of the input text to be used for modifying the LLM-based translations. However, all these methods are static; that is, they do not change with respect to the variation in the mood, genre, or context, which is an inherent property of the entertainment content. Similarly, \citet{Li_Chen_Yuan_Wu_Yang_Tao_Xiao_2024} tries to induce cultural nuances of the target language by introducing a knowledge base (KB) for idioms, which are difficult to translate in general. However, these models do not cover Indian languages, which have their own structural and lexical nuances \cite{leong2023bhasaholisticsoutheastasian}.

\noindent{\bfseries{LLMs for Entertainment Translation:}} Machine Translation using LLMs has started to gain popularity in recent times \cite{brown2020language, promptingllm, tao2024catllm}. Broadly, this can be classified into two categories: (i) prompt-based guiding and (ii) translation memory/RAG-based translation aiding. Below, we point out the issues with these techniques when applied to entertainment translation.

\begin{figure*}[h]
    \centering
    \includegraphics[width=0.85\linewidth]{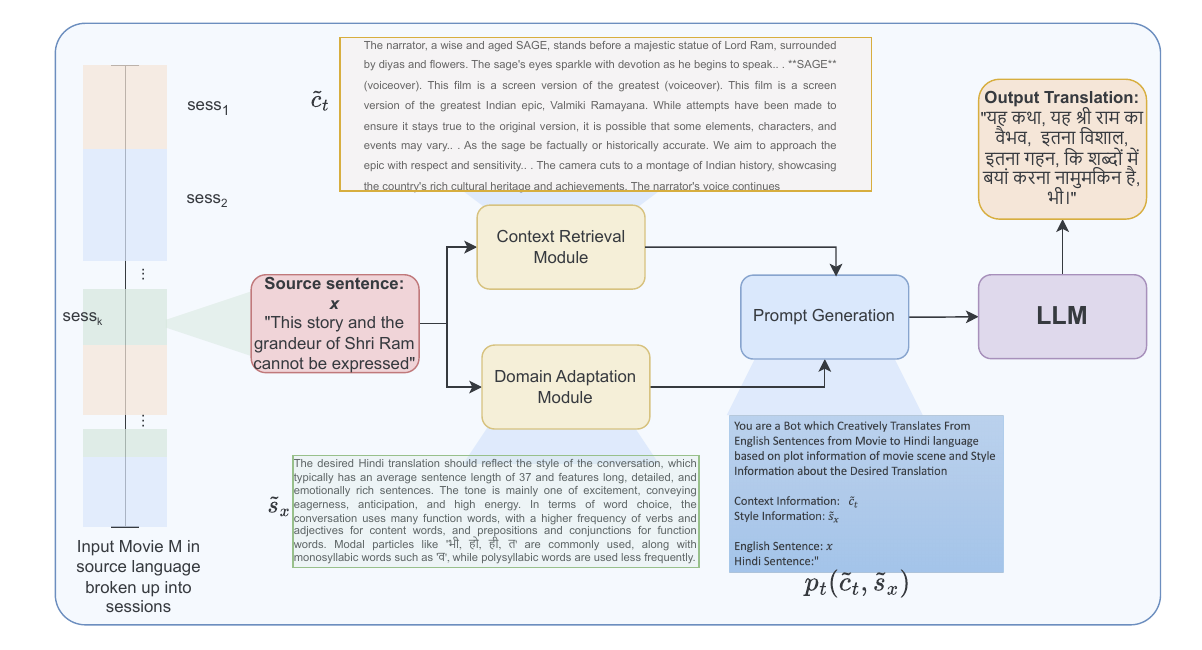}
    \caption{A high-level overview of our proposed methodology.}
    \label{fig:main-block}
\end{figure*}

\begin{itemize}
    \item [(i)] {\bfseries Prompt-based Guiding:} Prompt-based guiding of LLMs to perform translation can be treated as providing a conditioning parameter $p$, \textit{viz}., the \textit{prompt}, to the translation model:
    \begin{equation*}
        P_{\theta}\left(y| x,p\right)=\prod\limits_{i=1}^{L}P_{\theta}\left(y_i|p,x,y_1,\ldots,y_{i-1} \right).
    \end{equation*}
    where $L$ is the length of the output sentence $y$. However, when working in the automatic dubbing application for movies and OTT content, the prompt needs to be time-dependent, i.e. $p\to p_t$, in order to deal with the dynamic context $c_t$. In particular, the prompt can be formulated as $p_t=h(p,c_t)$, where $h$ is a linking and weight function in a latent space. The adaptive nature of the prompt $p_t$ induced by the time-varying context $c_t$ is vital in generating context-relevant translations for dubbing applications. However, it has received limited attention from researchers \cite{gao-etal-2023-adaptive}.

    \item [(ii)] {\bfseries Translation memory-based approach:} Traditional retrieval-aided translation systems have two primary components: (i) a retriever $p_{\eta}(.|x)$ which gives a probability distribution over a set of hidden context vectors stored in a vector database, and (ii) a generator $p_w(.|x,z)$ which gives a probability distribution over the output tokens given the source sentence $x$ and context $z$. The retriever aims at providing additional information to the generator, which is an LLM performing translation, by retrieving context $z$ by Maximum Inner-Product Search (MIPS) \cite{RAG}. However, the retrieved context vectors $z$ are semantically similar to the query sentence $x$ and do not take into account the style $s_x$ of the source sentence, for example, politeness, (in-)formality, regional dialect, etc. \cite{tao2024catllm}

\end{itemize}

\subsection*{Potential Resolution}\label{subsec:potential reolution}
The above-mentioned limitations reflect the need for a machine translation system that takes into account the context $c_t$ and preserves the style $s_x$ of a given source input sentence $x$. To this extent, a potential solution is to segment the (sequential) text into \textit{sessions}, where the `genre' of the sentences in a session remains constant. 
These `constant mood' sessions can be used to estimate the context and style, i.e., $\Tilde{c}_t$ and $\Tilde{s}_x$.
By incorporating this additional information, a time-varying prompt $p_t(\Tilde{c}_t, \Tilde{s}_x)$ can be obtained to leverage LLM's reasoning and understanding capabilities for generating context and style-aware translations. 

\begin{algorithm}[ht!]
\caption{Genre Classification and Segmentation}\label{alg:genre_segmentation}
    \resizebox{\linewidth}{!}{
        \begin{minipage}{\linewidth}
            \begin{algorithmic}[1]
                \State \textbf{Input:} $\cM$, clusters of the three classes, minimum number of sentences for a new session $(\alpha)$, maximum number of sentences in a session $(\beta)$
                \State Extract embeddings for each $x\in \cM$ and use $k$-NN to assign its class label and store in a array $g$.
                \State current-session $\gets$ \{$g[0]$\}
                \State session-list $\gets$ $\emptyset$
                \While{$i < $ length($g$)}
                    \If{length(current-genre) == $\beta$}
                        \State session-list $\gets$ current-session
                        \State current-session $\gets$  $\emptyset$
                    \EndIf
                    \If{$g[i]$ $\neq$ current-session[0] $\land$ length(current-session) $\geq \alpha$}
                        \State majority-label $\gets$ MAJORITY($g[i]:g[i+\alpha]$)
                        \If {majority-label $\neq$ current-session[0]}
                            \State session-list $\gets$ current-session
                            \State current-session $\gets$ $\emptyset$
                        \EndIf
                    \EndIf
                    \State current-session $\gets$ $g[i]$
                    \State $i \gets i + 1$
                \EndWhile
                \State\textbf{Output:} session-list
            \end{algorithmic}
        \end{minipage}
    }
\end{algorithm}

\section{Methodology}\label{sec:methodology}
In this section, we describe our methodology beginning with stating the problem statement formally. Next, we explain the necessity of segmenting the input text and how to obtain it. We then describe the method for extracting the $\Tilde{c}_t$ and $\Tilde{s}_x$ for a particular dialogue $x$ to generate the context and style-aware prompt $p_t$.

\subsection{Problem Formulation}\label{sec:problem formulation}
We consider the entertainment translation as an extension of neural machine translation task, where we primarily try to translate sentences from a source language ($\cS$) to a target language ($\cT$). These sentences can be dialogues from movies, web series, novels, etc. Formally, let $\mathcal{D}^{\cS}$ be defined as the set of all sentences in a $\cS$ and $\mathcal{D}^{\cT}$ the corresponding set in $\cT$. The goal of a translation system is to find a mapping $g:D^{\cS}\mapsto\cD^{\cT}$. However, translating movie dialogues from one language to another requires additional knowledge of the running \textit{context} $c_t$ as well as the style $s_x$ of the source sentence $x$. Hence, we define a mapping $g_E$, which is specific to translation in the entertainment domain, as a function that outputs the translated text $y$ as $y=g_E(x;c_t,s_x)$.
In other words, the aim of an entertainment translation system is to find a mapping $g_E$ which not only translates any  $x\in \mathcal{D}^{\cS}$, into a sentence $y\in \mathcal{D}^{\cT}$ but also preserves the context $c_t$ and the style $s_x$ of the input source sentence. Further, the mapping learned should be such that it induces creativity in the translation, which can increase the target audience's interest and engagement. These additional factors make the task of entertainment translation unique and challenging.

\subsection{Adaptive Session Classification and Segmentation}
For this section, we will take a concrete example of a movie $\cM$ to explain the key concepts (note: we only consider the sequence of text dialogues as $\cM$).
In entertainment content, each movie or web series is characterized by a sequence of scenes, each belonging to a specific \textit{genre} or tone, such as action, horror, comedy, and so forth. 
 Therefore, a movie $\cM$ can be represented as $\cM  = (sess_1, sess_2,\ldots, sess_M)$, where $M$ is the total number of scenes/sessions in the movie. Suppose a dialogue $x \in sess_k$, the style of translation of $x$ is expected to be more likely dependent on the current and $K$ neighboring sessions than much older sessions, which necessitates the segmentation of the text to ensure translation quality.

We provide an offline algorithm for achieving adaptive segmentation of $\cM$ in Alg. ~\ref{alg:genre_segmentation}, which classifies each session into one of the three three primary tonal categories: Serious (Intense genres: action, mystery, thriller, horror), Casual (Light genres: comedy, romance, fantasy) and Neutral (Dialogues with low emotional intensity). While not all the input texts may fit perfectly into these three categories, this approach provides a foundation for simple yet consistent classification by grouping genres with similar tones. We pretrain a $k$-NN classifier and generate clusters using example dialogues from the three categories. We refer the reader to the Appendix for details on the segmentation process. We also remark that Alg.~\ref{alg:genre_segmentation} only provides a rough estimate of the session boundaries of $\cM$. Next, we demonstrate how we extract the context and style information from the available sessions.

\subsection{Session Information Generation}
This section provides a thorough insight to the crux of our method. Let the current input dialogue be $x$. As depicted in Figure ~\ref{fig:main-block}, $x$ passes through two separate pipelines for $\Tilde{c}_t$ and $\Tilde{s}_x$ extraction. Subsequent paragraphs provide detailed description of these blocks.
\begin{figure}[h]
    \centering
    \includegraphics[width=\linewidth]{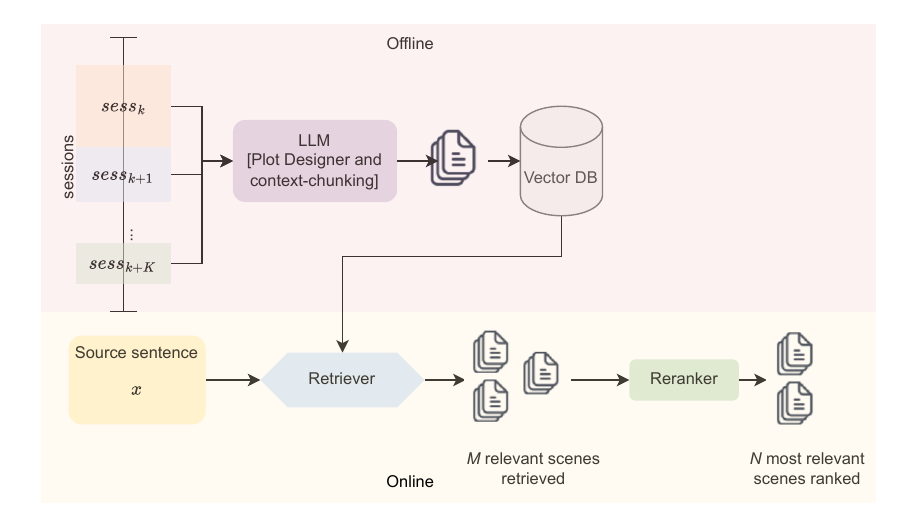}
    \caption{A block diagram of the Context retriever block.}
    \label{fig:rag-paper}
\end{figure}
\subsubsection{Context retrieval--Advanced RAG:}\label{subsubsec:context retriever}
Using Large Language Models (LLMs) to translate dialogues from one language to another without any prior context can lead to \textit{disconnected} translations, especially in a conversation. In order to induce interest among the target audience, LLMs can generate creative translations, which may lead to hallucinations \cite{zhang2023sirenssongaiocean}. Hence, providing the current session information can guide the LLM in translating the source sentence creatively with respect to the context of the movie, reducing hallucinations. 

As depicted in Figure ~\ref{fig:rag-paper}, we consider an offline process to extract the \textit{plots}, \textit{i.e.}, a summary of movie scenes, from $K$ consecutive sessions via an LLM. This extracted context is then subdivided into small chunks and stored in a vector database. This chunking helps our methodology two-fold. Firstly, the generated prompt might be too large for the LLM to comprehend. Secondly, the most relevant chunk/scene for the source sentence could well be from a different session (in the past or in the future). During the translation phase, a retriever uses the source sentence $x$ to retrieve $M$ most relevant chunks from the vector database \cite{RAG}. This is then passed through a re-ranker \cite{glass2022re2gretrievererankgenerate}, to generate $N$ most relevant chunks in a ranked fashion, which we denote as $\Tilde{c}_t$ for sentence $x$.

\begin{figure}[h]
    \centering
    \includegraphics[width=1.1\linewidth]{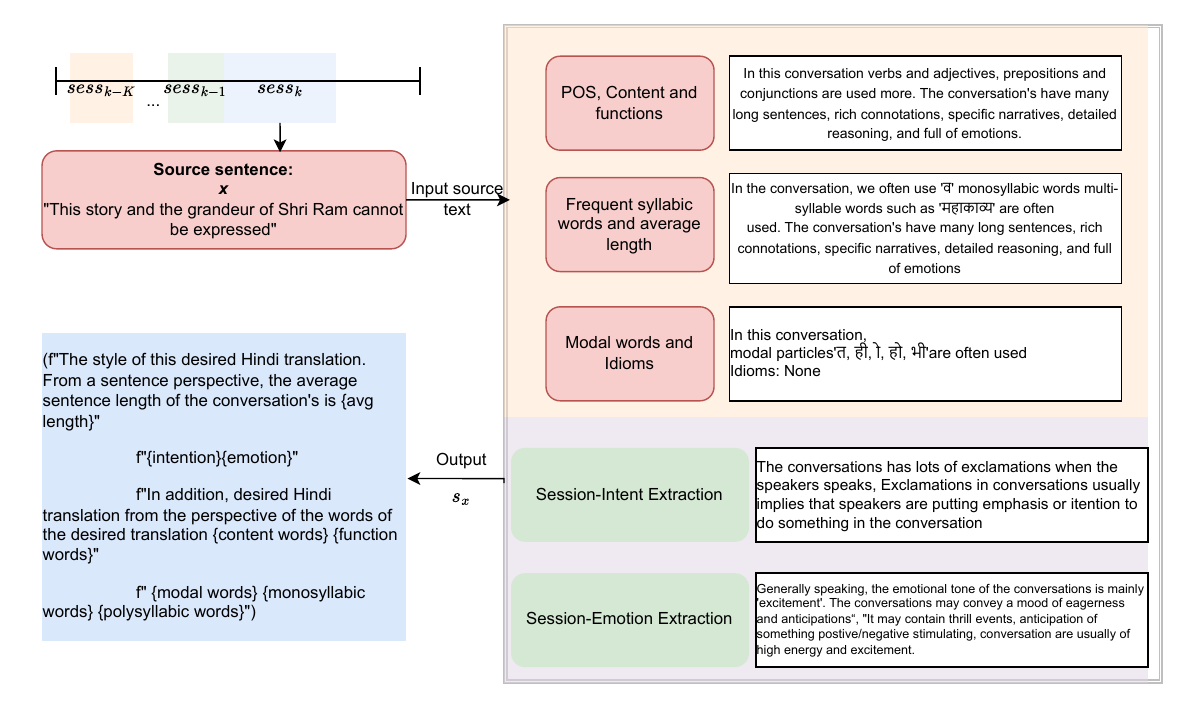}
    \caption{Domain Adaptation Module}
    \label{fig:DAM}
\end{figure}

\subsubsection{Style extraction-Domain Adaptation Module:}\label{subsubsec:style extractor}
By using the above pipeline for context information extraction, we can generate creative translation that aligns with the current context and mood of the scene. However, this does not help in extracting the style or tone of the dialogue. In particular, $\Tilde{c}_t$ does not include the most used words, idioms, and emotional state of the current scene, which define the overall language register. To tackle this, we designed a Domain Adaptation Module (DAM), which is a collection of various information-extracting NLP subroutines. These subroutines help in constructing $\Tilde{s}_x$, which acts as a clear and comprehensive style-determining prompt to be fed to the LLM. We note that we get inspired from \cite{tao2024catllm} with changes in the DAM module owing to our specific application and the change in language family. In particular, we pay special attention to dialogue and session-level information, respectively, which is in contrast with their approach, which dealt with style transfer as a one-shot method for the entire text at once. Subsequently, we explain these modules in detail.

\noindent{\bfseries{Dialogue Level Module}:}
This module provides the structural information of the dialogues, giving us the overall conversational style of speakers. It consists of three parts as described in brief below.
\begin{itemize}
    \item {\bfseries Content and Function words}: Here, we take the output translations of the past $K$ sessions as input and pass it through a PoS Tagger trained on Indic languages. We categorize these tagged words into \textit{content words} and \emph{function words}\cite{carnap1967logical}, which we then convert to the respective prompts $f_c$ and $f_f$. For the explicit prompts, we refer the reader to the Appendix.

    \item {\bfseries Frequent Syllabic Words:} Every speaker may have a different style of speaking for instance, depending on the regional dialect, pronouns like "I" or "myself" can be termed in Hindi as ``\textit{apun}", (spoken in Mumbai region) or ``\textit{hum}", (spoken in northern India), etc. Identifying this will provide the model with information on the frequent use of \textit{monosyllabic}  and \textit{polysyllabic} words. Similar to the above case, we convert them into prompts as $f_m$ and $f_p$ respectively.
    
    \item {\bfseries Modal Words and Idioms:} Modal words and idioms contribute to the tone, politeness, and effectiveness of the conversation ($f_{modal}$, $f_{idioms}$ respectively). 
\end{itemize}

\noindent{\bfseries{Session Level Module:}}
In contrast with the dialogue-level information extraction, the session-level module allows an understanding of the global intent of the ongoing and past sessions.
\begin{itemize}
    \item {\bfseries Sentence Intent and Emotion:} Intent of a session can be derived from the use of punctuation marks. For instance, excessive use of question marks in a particular scene can indicate the scene to be interrogatory. Hence, we count all the punctuation in session, then define intent based on thresholds ($f_{intent}$). Further, to extract the emotion, we pass the current session through an LLM to generate $f_{emotion}$. Furthermore, Figure \ref{fig:DAM} depicts details of the domain adaptation module with a concrete example.
\end{itemize}

\noindent Finally, we obtain the Context and Style Aware prompt $p_t$, by concatenating the outputs from the context retrieval module ($\Tilde{c}_t$) and the DAM module ($\Tilde{s}_x$). We refer the reader to the Appendix, where we
illustrate detailed examples of prompt $p_t$ for enhanced clarity.

\begin{algorithm}[ht!]
\caption{Context and Style Aware Translation (CASAT)}\label{alg:main pipeline]}
    \resizebox{\linewidth}{!}{
        \begin{minipage}{\linewidth}
        \begin{algorithmic}[1]
        \State \textbf{Input:} Source Sentences $(\cM)$, $M$, $N$, session-list (See Alg.\ref{alg:genre_segmentation})
        \For{$x \in \cM$}
            \State  $\tilde{c}_t\gets$ Extract $M$ relevant scenes from the vector DB and choose $N$ best through the Context Retriever Module.
            \State $\tilde{s}_x\gets$ Extract dialogue level and session level information through DAM.
            \State $p_t \gets$ Generate prompt using $\tilde{c}_t$ and $\tilde{s}_x$
            \State Translation $y_x$ $\gets$ LLM($p_t,x$)
        \EndFor
        \end{algorithmic}
        \end{minipage}
}
\end{algorithm}

\begin{figure*}[t!]
     \centering
    \begin{subfigure}[b]{.5\textwidth}
         \centering
         \includegraphics[width=7cm, height=4cm]{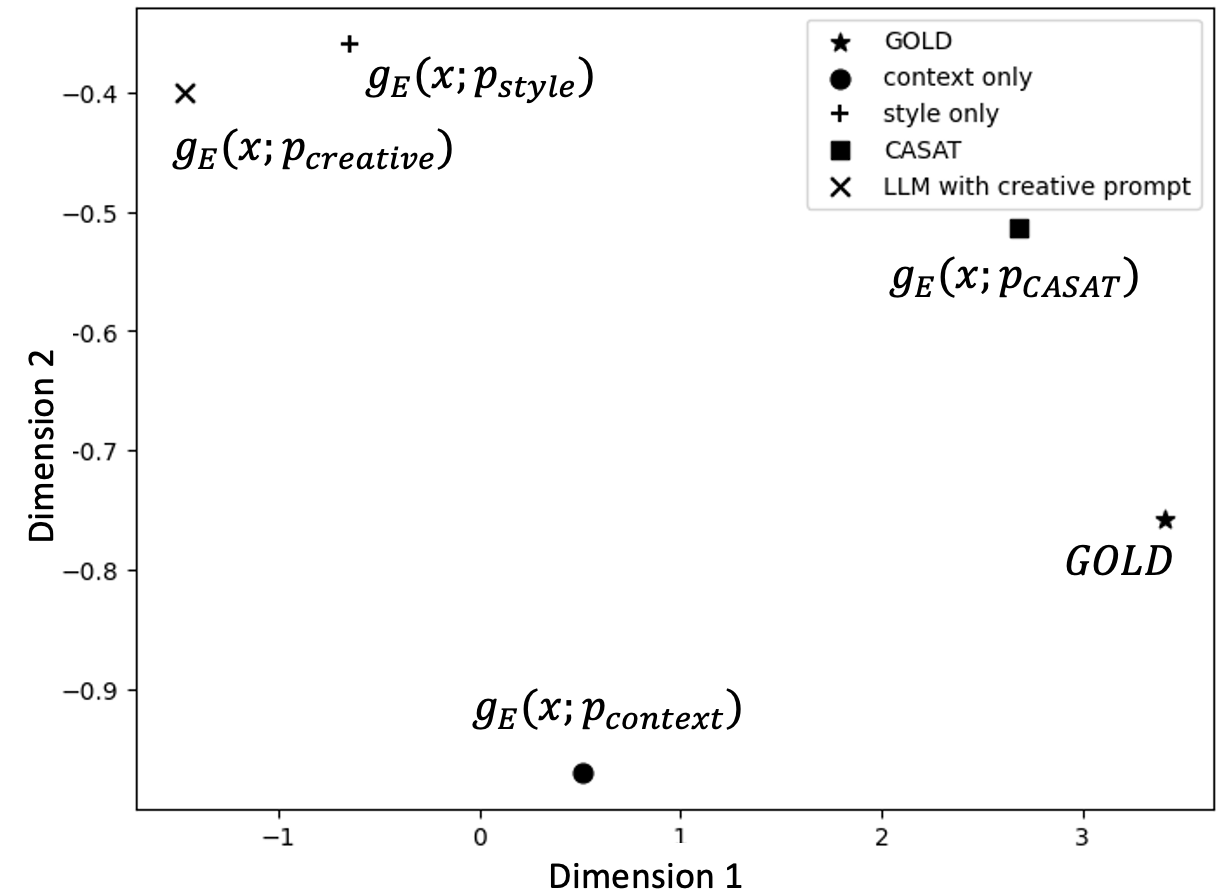}
         \caption{MDS plot of the translations generated via different prompts.}
         \label{fig:labelled_our_Approach}
     \end{subfigure}%
     \hfill
     \begin{subfigure}[b]{.5\textwidth}
         \centering
         \includegraphics[width=7cm, height=4cm]{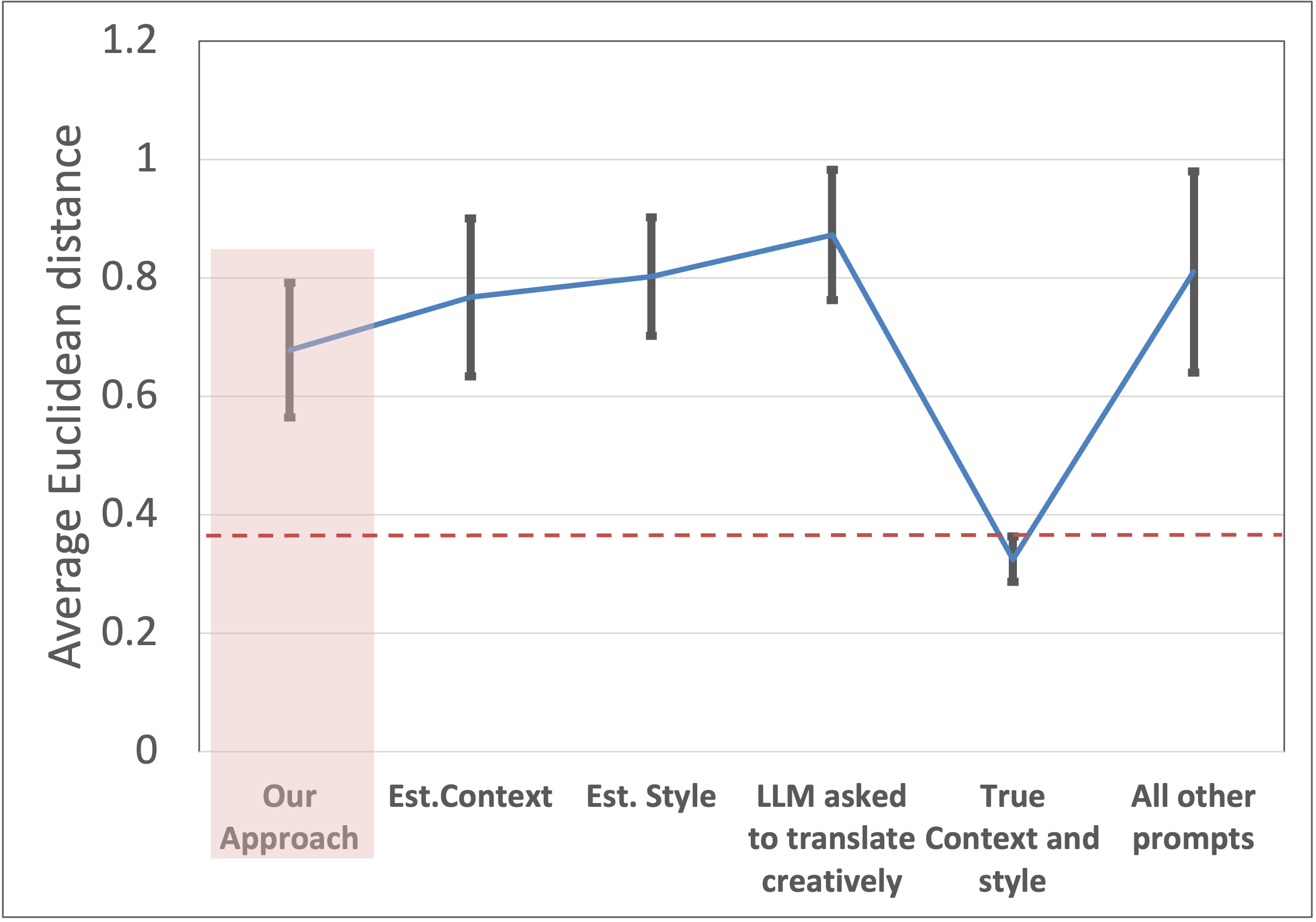}
         \caption{Embedding distances from the reference output}
         \label{fig:errorbar}
     \end{subfigure}
        \caption{A comparative analysis of the effect of various prompts on the translated text. }
        \label{fig:four graphs}
        
\end{figure*}

\begin{table*}[h]
    \centering
    \resizebox{\textwidth}{!}{
    \begin{tabular}{|c|c|c|c|c|c|c|c|c|c|c|c|c|c|c|c|c|}
        \hline
        \multirow{3}{*}{\textbf{LLMs Sizes}}& \multirow{3}{*}{\textbf{Models}} & \multicolumn{5}{|c|}{\textbf{En-Hi}}& \multicolumn{5}{|c|}{\textbf{En-Ben}}& \multicolumn{5}{|c|}{\textbf{En-Tel}}\\
        \cline{3-17}
        & & \multicolumn{2}{|c|}{Base}& \multicolumn{3}{|c|}{CASAT}& \multicolumn{2}{|c|}{Base}& \multicolumn{3}{|c|}{CASAT}& \multicolumn{2}{|c|}{Base}& \multicolumn{3}{|c|}{CASAT}\\
        \cline{3-17}
        & & B.& C.& B.& C.& $\Delta$& B.& C.& B.& C.& $\Delta$& B.& C.& B.& C.& $\Delta$\\
        \hline
        \multirow{4}{*}{\textbf{Small-Sized LLMs}}& Mistral 7B& 1.33& 0.41& 1.55& 0.42& 0.56& 0.2& 0.43& 0.38& 0.48& 0.62& 0.1& 0.42& 0.07& 0.41& 0.8\\
        \cline{2-17}
        & LLaMa-3 8B& 3.42&0.51 &4.51 &  0.56 &0.56&  0.75& 0.61 &1.12&  0.67&0.52&  0.85& 0.51 &0.60& 0.58&0.69 \\
        \cline{2-17}
        & Aya23 8B& 6.67&0.61&\textbf{7.21}&  \textbf{0.64}&0.67&  0.30& 0.48 &0.4&  0.53&0.56&  0.16& 0.42 &0.3& 0.46&0.76 \\
        \cline{2-17}
        & Gemma2 9B& 6.68&0.56 
 &6.88 &  0.62 &0.62&  1.55& 0.68 
&\textbf{2.53}&  \textbf{0.75}&0.62&  1.43& 0.65 
&\textbf{1.75}& \textbf{0.69}&0.79\\
        \hline
        \multirow{2}{*}{\textbf{Mid-Sized LLMs}}& Gemma2 27B& 4.71&0.62 
 &8.07 &  0.67&0.69&  1.49& 0.70 
&\textbf{3.08}&  \textbf{0.77}&0.67&  1.7& 0.67 
&\textbf{2.07}& \textbf{0.71}&0.77 \\
        \cline{2-17}
        & Aya23 35B& 9.25&0.63 
 &\textbf{9.59}&  \textbf{0.68}&0.70&  0.80& 0.62 
&0.82&  0.65&0.59&  0.17& 0.44 
&0.23& 0.48&0.8\\
        \hline
        \multirow{2}{*}{\textbf{Large-Sized LLMs}}& LLaMa3 70B& 7.96&0.63 
 &9.84&  0.70&\textbf{0.73}&  2.46& 0.66 
&2.09&  0.75&0.74&  0.92& 0.60 
&1.11& 0.65&0.85 \\
        \cline{2-17}
        & GPT-3.5 Turbo& 11.84&0.69  &\textbf{14.44} &  \textbf{0.72} &0.73&  12.88& 0.79&\textbf{14.91}&   \textbf{0.82}&0.75&  7.41& 0.66&\textbf{10.9}&  \textbf{0.78}&0.85\\
        \hline
    \end{tabular}
    }
    \caption{Performance comparison of CASAT with various SOTA LLMs fed with prompts to generate creative translations. Here B.:BLEU, C.:COMET score in range [0,1]}
    \label{tab:main table}
\end{table*}

\section{Experiments}\label{sec:experiment}
In this section, we present the experimental evaluation of our proposed approach. We will also describe the effect caused by the individual components of CASAT through ablation studies. All experiments were carried out on 1x1H100 80 GB GPU.
\subsection{Experimental Settings}
{\bfseries Evaluation Dataset:}
This section provides the necessary details of the datasets we used for evaluating Alg\ref{alg:main pipeline]}. Since our method does not have an explicit training phase, we describe the data that we use for testing Alg.\ref{alg:main pipeline]}. In addition, due to the unavailability of Indian language entertainment domain public datasets, we use web-scrape data for our simulations, which will be explained later.

We scrapped parallel text data of popular movies from the popular subtitle website \textit{OpenSubtitles.org}. To see the effect on text data, which requires even more human-induced creativity, we further used parallel text data from a popular children 's cartoon series. All our experiments are conducted on the set of three language directions, \textit{viz.}, English-to-Hindi (En-Hi), English-to-Bengali (En-Ben) and English-to-Telugu (En-Tel). Next, we mention the specific details of the text content used for our numerical studies. We label our text data into three categories: (i) literal, (ii) semi-creative, and (iii) creative, owing to the increasing levels of creativity in the reference gold data. 
\begin{itemize}
    \item {\bfseries English-to-Hindi:} We choose subtitles scrapped from the \textit{opensubtitles} website for the following movies: (i) \textit{Adipurush} (creative), (ii) \textit{Pushpa} (semi-creative), and (iii) \textit{Interstellar} (literal). In addition, we use episodes from a popular cartoon series (creative) for evaluation. The total number of sentence pairs for En-Hi was 5238.
    
    \item {\bfseries English-to-Bengali:} Similarly, we scrapped subtitles of two movies, namely \textit{Wolves} and \textit{Maharaja} from \textit{opensubtitles} website was used for evaluation for this language pair, amounting to a total of 3259 sentence pairs.

    \item {\bfseries English-to-Telugu:} For English to Telugu translation, we scrapped subtitles of two movies, namely \textit{Without Remorse} and \textit{Bumblebee} from \textit{opensubtitles} for evaluation, comprising of 1698 sentence/dialogue pairs.
\end{itemize}

\noindent{\bfseries LLMs used for comparison:} We randomly select 800 data samples from each language source and translate them to the target language utilizing 5 distinct Large Language Models with varying sizes, categorizing them into three sections:
\begin{itemize}
    \item {\bfseries Small Sized LLMs:} We focused on three multi-lingual small sized models which perform well in Indic Languages i.e Mistral 7B \cite{jiang2023mistral7b}, Gemma2 9B \cite{gemmateam2024gemma2improvingopen}, Aya23 8B \cite{aryabumi2024aya23openweight}, Llama3 8B \cite{dubey2024llama3herdmodels}.

    \item {\bfseries Mid-sized LLMs:} We consider two LLMs, namely, Gemma2 27B \cite{gemmateam2024gemma2improvingopen} and Aya23 35B \cite{aryabumi2024aya23openweight}, for mid-sized category. Both of these LLMs have performed consistently well in Indic languages.

    \item {\bfseries Large-sized LLMs:} Likewise we considered two large-sized LLMs that are Llama3 70B \cite{dubey2024llama3herdmodels} and GPT-3.5 Turbo, both having excellent reasoning and translation qualities.
\end{itemize}

\begin{table*}[!ht]
    \centering
    \resizebox{\textwidth}{!}{ 
    \begin{tabular}{|c|l|l|c|c|l|c|c|l|c|c|l|} \hline 
         \multirow{2}{*}{\textbf{Model}}& \multicolumn{2}{|c|}{\textbf{Baseline}}&  \multicolumn{3}{|c|}{\textbf{Context Only}}&  \multicolumn{3}{|c|}{\textbf{DAM Only}}&  \multicolumn{3}{|c|}{\textbf{CASAT}}\\ \cline{2-12}
          & BLEU&COMET&  BLEU&  COMET & 
 $\Delta$&  BLEU&  COMET &$\Delta$&  BLEU& COMET &$\Delta$\\ \hline 
 Mistral 7B& 1.33& 0.41& 1.16& 0.41&0.67& \textbf{1.89}& 0.41&0.67& 1.55&\textbf{0.42}&0.56\\ \hline 
 Llama3 8B& 3.42& 0.50& 3.83& 0.55&0.58& 3.43& 0.51&0.55& \textbf{4.51}&\textbf{0.56}&0.60\\ \hline 
 Aya23 8B& 6.67& 0.61& 6.77& \textbf{0.64}&0.66& 7.01& 0.62&0.64& \textbf{7.21}&0.64&0.67\\ \hline 
 Gemma2 9B& 6.68& 0.56& \textbf{7.8}& \textbf{0.67}&0.61& 7.14& 0.59 &0.60& 6.88&0.62 &0.62\\ \hline 
 Gemma2 27B & 4.71& 0.62& 7.46& 0.66&0.62& 5.17& 0.64&0.66& \textbf{8.07}&\textbf{0.67} &0.69\\ \hline 
 Aya23 35B & 9.25& 0.63& 6.95& 0.67&0.67& 8.32& 0.66&0.70& \textbf{9.59}&\textbf{0.68} &0.70\\ \hline 
 Llama3 70B& 7.96& 0.62& 7.12& 0.66&0.71& 8.99& 0.67&0.64& \textbf{9.84}&\textbf{0.70}&0.73\\ \hline
    \end{tabular}}
    \caption{Analysis of the effect of the individual components of CASAT}
    \label{tab:ablation}
\end{table*}

\noindent{\bfseries Evaluation Metrics:} We adopt three metrics for the evaluation task. SacreBLEU \cite{sacrebleu} represents $n$-gram matching while COMET (wmt22-cometkiwi-da) \cite{rei-etal-2022-cometkiwi} represents the reference-free neural-based evaluation. Third, we use GPT-4o for evaluation of the translated text, which is well-known to replicate human-level judgment \cite{fu2023gptscoreevaluatedesire} by calculating the win-ratio $(\Delta)$ of our approach over the baseline models as follows:
\begin{equation*}
\small
    \Delta=\frac{\left(\begin{split}\#\text{times GPT-4o chooses CASAT based} \\ \text{translation over baseline LLM translation}\end{split}\right)}{\# \text{total translations}}.
\end{equation*}

\subsection{Can CASAT provide audience-engaging translations?} 

\noindent{\bfseries Main Result and Analysis.} The outcomes presented in Table\ref{tab:main table} illustrate that our method demonstrates superior performance by consistently incorporating plot and style information compared to directly prompting creativity in LLMs (see the exact prompt used for baseline LLMs in Appendix). Secondly, irrespective of the LLM chosen to produce the translation, CASAT significantly enhances its quality across the evaluation metrics. Interestingly, Mistral 7B shows minimal enhancement for En-Hi and En-Tel directions, yet it exhibits a commendable win ratio for En-Ben and En-Tel directions. Thirdly, both the performance in win-ratio \textit{and} COMET scores improve with larger model sizes, suggesting that increasing the model size enhances LLM's capability of plot development and comprehension of the in-context information.
However, surprisingly we observe that the 9B and 27B versions of Gemma2 either perform similarly to or even outperform models such as Aya23 35B and Llama3 70B for En-Ben and En-Tel language directions in terms of COMET scores.

\noindent{\bfseries How does the inclusion of context and style impact the resulting output? } We plot the multi-dimensional scaling (MDS) representation of the generated text from Llama 3-8B, with varying prompts in Figure\ref{fig:labelled_our_Approach}. We observe that prompting the LLM differently affects the output translation in a significant manner, as also reported in \cite{salinas2024butterflyeffectalteringprompts}. The plot indicates that solely incorporating the style has minimal impact on the translation quality, whereas solely providing the plot information (context) enhances the quality, evident by the reduced distance between the context and reference in comparison to style alone. CASAT, i.e., the simultaneous provision of context and style, significantly enhances the quality of the translation. Figure\ref{fig:errorbar} plots the average Euclidean distance of the generated text from the reference translations for a range of prompts. The plot shows that CASAT is closest to the reference translation.

\noindent{\bfseries How does CASAT Fare Against Traditional MT Systems?} We evaluate the Win-Ratio ($\Delta$) of CASAT-augmented models against traditional machine translation (MT) systems across En-Hi, En-Ben, and En-Tel translation directions. Specifically, we compare the performance of Gemma2 9B (CG9) and Gemma2 27B (CG27) models, enhanced with the CASAT approach, against traditional systems such as IndicTrans2 (ITv2) \cite{gala2023indictrans} and NLLB \cite{nllbteam2022languageleftbehindscaling}. The results, summarized in Table \ref{tab:CSVSTMT}, demonstrate that CASAT-augmented models are consistently preferred in the entertainment domain, underscoring the effectiveness of the CASAT approach in improving translation quality, particularly in domain-specific contexts.  

\begin{table}
    \centering
    \begin{tabular}{|c|c|c|c|} \hline 
         \textbf{Models}&  \textbf{En-Hi}&  \textbf{En-Ben}& \textbf{En-Tel}\\ \hline 
         CG9 vs ITv2 &  58\% &  51\% & 53\% \\ \hline 
         CG27 vs ITv2 &  65\% &  58\% & 53\% \\ \hline 
         CG9 vs NLLB &  66\% &  51\% & 54\% \\ \hline 
 CG27 vs NLLB & 64\% & 61\% &68\% \\ \hline
    \end{tabular}
    \caption{Wini-ratio of CASAT- vs Traditional MT Systems}
    \label{tab:CSVSTMT}
\end{table}

\noindent{\bfseries How many sessions $K$ to consider?} The performance of all models on En-Hi language pair datasets are compared for various values of $K$ in {Table \ref{tab:value of K varying}}. Since $K$ is utilized in plot design and DAM, it is a crucial parameter to consider. Generally, it has been observed that $K=2$ and $K=3$ exhibit good performance. The results indicate that using $K=1$ yields insufficient contextual information, while K=4 results in less specificity. 

\begin{table}[!ht]
    \centering
    \begin{tabular}{|c|c|c|c|c|} \hline 
         \textbf{Models}&  \textbf{K=1}&  \textbf{K=2}&  \textbf{K=3}& \textbf{K=4}\\ \hline 
 Mistral 7B& 0.367& \textbf{0.424}& 0.402&0.371\\ \hline 
 Llama3 8B& 0.487& \textbf{0.562}& 0.534&0.507\\ \hline 
 Gemma2 9B& 0.644& 0.62& 0.647&\textbf{0.658}\\ \hline 
 Aya23 8B& 0.637& 0.64& \textbf{0.65}&0.644\\ \hline 
 Gemma2 27B& 0.66& \textbf{0.67}& 0.64&0.63\\ \hline 
 Aya23 35B& 0.67& 0.68& \textbf{0.69}&0.66\\ \hline 
 Llama3 70B& 0.68& \textbf{0.70}& 0.67&0.66\\ \hline
    \end{tabular}
    \caption{COMET scores showing the effect of varying the value of number of sessions $K$ .}
    \label{tab:value of K varying}
\end{table} 

\subsection{Ablation Studies}
We conduct ablation studies on the effect of the domain adaptation module for style transfer and the context retriever block and compare the results with the respective baseline LLMs. We show the BLEU scores, COMET scores, and win-ratios in Table \ref{tab:ablation} for all the considered LLMs. We observe that providing `context only' improves the relevancy of output translation, which is reflected in COMET and win ratio scores. On the other hand, `DAM only' helps to navigate the output to copy the style of text and hence a larger value for the metric BLEU. Finally, combining the two, i.e., for CASAT, we obtain better BLEU score, COMET, and win-ratios across LLMs, which we conjecture that the LLM is able to gain complementary information from each of the two blocks.

\section{Conclusion}\label{sec:conclusion}
We explored the challenging task of entertainment translation, where we identified two key aspects, context, and style, which make this problem unique. We proposed a methodology to estimate these factors and use them to generate context and style-aware translations from an LLM. We showcased the efficacy of our algorithm via numerous experiments using three Indian language entertainment text datasets and various LLMs. Important future directions include using sophisticated methods for automatic segmentation of text, such as text diarization. Further, our approach has an offline component for partitioning of sessions and generation of contextual information, which we intend to eliminate to develop a completely online algorithm. 

\bibliography{aaai25}

\begin{table*}[htb!]
    \centering
    \begin{tabular}{|c|c|c|c|c|l|l|} \hline 
         \textbf{Model}&  \multicolumn{2}{|c|}{\textbf{Past $l$ as Context}}&  \multicolumn{2}{|c|}{\textbf{Past $l/2$ and Next $l/2$ as Context}} & \multicolumn{2}{|c|}{\textbf{Context Retrieval- Advanced RAG}}\\ \hline 
         &  BLEU&  COMET&  BLEU& COMET& BLEU&COMET \\ \hline 
         Aya23 8B&  6.38&  0.61&  6.42& 0.61& \textbf{6.77}&\textbf{0.64 }\\ \hline 
         Gemma2 9B&  6.56&  0.63&  6.50& 0.62& \textbf{7.8}&\textbf{0.67} \\ \hline 
         Gemma2 27B&  7.21&  0.65&  7.25& \textbf{0.66}& \textbf{7.46}&\textbf{0.66} \\ \hline 
         Aya23 35B&  6.4&  0.63&  6.7& 0.66& \textbf{6.95}&\textbf{0.67} \\ \hline
    \end{tabular}
    \caption{Comparison between passing our proposed context against passing surrounding $l=10$ sentences as context. }
    \label{tab:SurrVSCntx}
\end{table*}

\newpage
\appendix
\section{Appendix}
\section{Additional Ablation Experiments}

\noindent{\bfseries Why use context-retrieval-Advanced RAG instead of surrounding sentences as context?} In the entertainment domain, particularly in movie dialogues, the plot of a scene (serving as context) often provides nuanced and crucial information for accurately translating the current dialogue, surpassing the utility of merely using adjacent sentences. As detailed in Section \ref{subsubsec:context retriever}, our proposed Retrieval-Augmented Generation (RAG) framework enables the retrieval of contextual information or plot details that may extend beyond the immediate neighboring scenes. This approach effectively captures dependencies with earlier or later scenes, which are critical for maintaining continuity and relevance in translation. When no such contextually significant scenes exist, RAG defaults to retrieving information from the current scene.
To validate the effectiveness of this approach, we conducted an experiment comparing two methods: (1) using Context Retrieval- Advanced RAG, and (2) using a fixed number $l$ of surrounding sentences as context. The surrounding sentences were defined as either (a) the past $l$ consecutive dialogues or (b) the past $l/2$ and next $l/2$ consecutive dialogues. This evaluation was performed on the English-Hindi translation direction using two sets of large language models (LLMs) of varying sizes. The results, presented in Table \ref{tab:SurrVSCntx}, demonstrate that incorporating plot information enhances the translation quality, as evidenced by improvements in both BLEU and COMET scores. These findings underscore the importance of leveraging plot-level context to achieve more contextually aware and coherent translations.

\section{Choice of hyper-parameters.}

\begin{table}[ht!]
    \centering
    \begin{tabular}{|c|c|} \hline 
         \textbf{Parameter}& \textbf{Value}\\ \hline 
         $K$& 2\\ \hline 
         $M$& 5\\ \hline 
         $N$& 2\\ \hline 
         chunk size& 356\\ \hline 
 chunk overlap&64\\ \hline 
 temperature for plot design&0.5\\ \hline 
 temperature for emotion and translation&0.2\\ \hline
 $\alpha$&5\\\hline
 $\beta$&10\\\hline
 $k$&3\\\hline
    \end{tabular}
    \caption{Values of the hyper-parameters chosen for our methodology.}
    \label{tab:hyperprametrs}
\end{table}

\section{Additional details of adaptive segmentation}
We synthetically generate 250 sample sentences of each genre, and store their vector representation $se$, $ne$, $ce$ respectively ( please refer to the plot below ). Next we loop over all the  $x \in \mathcal{D}^{\cS}$ and classify them shown below
\begin{equation*}
f(x) = \texttt{cosine-similarity}(\texttt{BERT}(x), (se,ne,ce))
\end{equation*}
\begin{equation*}
\texttt{genre}(x) = {\operatorname{argmax}} f(x)
\end{equation*}
Once all the sentences are classified based on genre we group them together using a window based grouping, where the maximum number of sentences in one group can be $\beta$ and minimum number of same genre to be present to classify as a new group in $\alpha$. In this way, we were able to generate several sequential sessions $S_x = \{S_t{}_1,S_t{}_2,..,S_t{}_n\}$, where $P_t{}_i$ had grouped sequential sentences based on similar genres. 
\begin{figure*}
    \centering
    \includegraphics[width=0.5\linewidth]{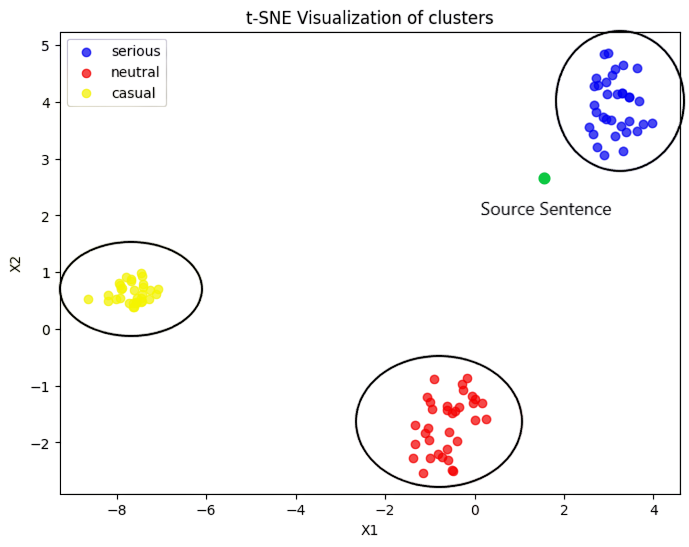}
    \caption{Clusters obtained using k-NN for genre segmentation. }
    \label{fig:genre-segmentation}
\end{figure*}

\section{Additional details of DAM}
\begin{equation*}
\texttt{tagged-words} = \texttt{POS-tagger}(sessions)
\end{equation*}
\begin{equation*}
f_c,f_f = \operatorname{argTopK} \{cf\_map(\texttt{tagged-words})\}
\end{equation*}
Once the pos tagger tags the words in the session it is passed through $cf\_map$ which maps the content and function words and provides the top K most occuring respectively
\section{Additional details on Experiments}

\noindent{\bfseries Possible Reasons for Low Automatic metric scores} The modern entertainment domain is highly nuanced, characterized by various ways to express the same idea. This results in a one-to-many mapping in translations, where the output can vary significantly based on the scriptwriter's creativity and cultural subtleties. Consequently, translations in this domain are rarely literal or word-for-word; instead, they prioritize conveying the intended meaning of the original dialogue while ensuring the translation remains engaging and relevant for native audiences.
This inherent variability is particularly pronounced in our challenging evaluation dataset, which features highly creative and culturally rich dialogues. The dataset includes examples where literal translations are not suitable, and preserving the context, tone, and cultural resonance is critical. Such content presents significant challenges for even the most advanced models, including GPT-3.5, often resulting in lower BLEU scores. To provide insight into the nature of this task, we include a few representative examples from the dataset in Figure \ref{fig:sampleD}.

\begin{figure*}
    \centering
    \includegraphics[width=1\linewidth]{samples-2.pdf}
    \caption{Samples of Source and Reference of the dataset}
    \label{fig:sampleD}
\end{figure*}

\begin{figure*}[h!]
    \centering
    \begin{subfigure}[b]{0.45\linewidth} 
        \centering
        \includegraphics[width=\linewidth]{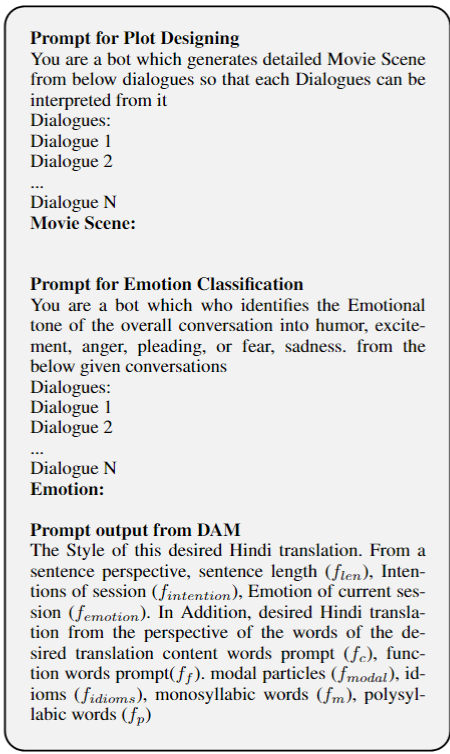}
    \end{subfigure}
    \hfill
    \begin{subfigure}[b]{0.45\linewidth} 
        \centering
        \includegraphics[width=\linewidth]{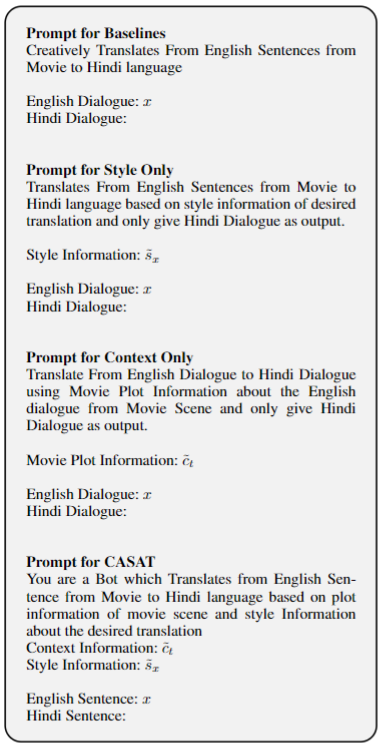}
    \end{subfigure}
    \caption{Explicit prompts used for the plot designer, session emotion extractor, and the final translation generator LLMs.}
    \label{fig:prompt}
\end{figure*}

\begin{table*}[h]
    \centering
    \resizebox{\textwidth}{!}{
    \begin{tabular}{|c|c|c|c|c|c|c|c|c|c|c|c|c|}
        \hline
        \multirow{2}{*}{\textbf{Model Size}}& \multirow{2}{*}{\textbf{Models}}& \multicolumn{2}{c}{\textbf{Base}}& \multicolumn{3}{c}{\textbf{Context Only}}& \multicolumn{3}{c}{\textbf{DAM Only}}& \multicolumn{3}{c}{\textbf{CASAT}}\\ \cline{3-13}
                                   &                           & BLEU& COMET& BLEU& COMET& $\Delta$& BLEU& COMET& $\Delta$& BLEU& COMET& $\Delta$\\ \hline
        \multirow{4}{*}{\textbf{Small Sized LLMs}}& Mistral 7B& 1.33& 0.41& 1.16& 0.41& 0.67& 1.89& 0.41& 0.67& 1.55& 0.42& 0.56\\ \cline{2-13}
                                   &                           LLaMa3 8B& 3.42& 0.50& 3.82& 0.55& 0.58& 3.43& 0.51& 0.55& 4.51& 0.56& 0.60\\ \cline{2-13}
                                   &                           Aya23 8B& 6.67& 0.61& 6.77& 0.64& 0.66& 7.01& 0.62& 0.64& 7.21& 0.64& 0.67\\ \cline{2-13}
                                   &                           Gemma2 9B& 6.68& 0.56& \textbf{7.80}& \textbf{0.67}& 0.61& 7.14& 0.59& 0.60& 6.88& 0.62& 0.62\\ \hline
        \multirow{2}{*}{\textbf{Mid-Sized LLMs}}&                           Gemma2 27B& 4.71& 0.62& 7.46& 0.66& 0.62& 5.17& 0.64& 0.66& 8.07& 0.67& 0.69\\ \cline{2-13}
                                   &                           Aya23 35B& 9.25& 0.63& 6.95& 0.67& 0.67& 8.32& 0.66& 0.70& \textbf{9.59}& \textbf{0.68}& 0.70\\ \hline
        \textbf{Large Sized LLMs}& LLaMa3 70B& 7.96& 0.62& 7.12& 0.66& 0.71& 8.99& 0.67& 0.64& \textbf{9.84}& \textbf{0.70}& 0.73\\ \hline
    \end{tabular}
    }
    \caption{Placeholder caption for a 13-column, 9-row table with merged cells.}
    \label{tab:placeholder_label}
\end{table*}

\end{document}